\documentclass[sigconf]{acmart}

\DeclareUnicodeCharacter{FF0C}{,}

\AtBeginDocument{%
  \providecommand\BibTeX{{%
    \normalfont B\kern-0.5em{\scshape i\kern-0.25em b}\kern-0.8em\TeX}}}

\usepackage{booktabs}
\usepackage{multirow}
\usepackage{threeparttable}
\usepackage{comment}
\usepackage{subcaption}
\usepackage{xspace}
\usepackage{amsmath}

\begin{document}

\title{3DFacePolicy: Audio-Driven 3D Facial Animation Based on Action Control}



\author{Xuanmeng Sha}
\affiliation{%
  \institution{The University of Osaka}
  \country{Japan}
}
\email{shaxuanmeng@gmail.com}

\author{Liyun Zhang}
\affiliation{%
  \institution{The University of Osaka}
  \country{Japan}
}
\email{liyun.zhang@lab.ime.cmc.osaka-u.ac.jp}

\author{Tomohiro Mashita}
\affiliation{%
  \institution{Osaka Electro-Communication University}
  \country{Japan}
}
\email{mashita@osakac.ac.jp}

\author{Naoya Chiba}
\affiliation{%
  \institution{The University of Osaka}
  \country{Japan}
}
\email{chiba@nchiba.net}

\author{Yuki Uranishi}
\affiliation{%
  \institution{Osaka University}
  \country{Japan}
}
\email{yuki.uranishi.cmc@osaka-u.ac.jp}



\begin{abstract}
  Audio-driven 3D facial animation has achieved significant progress in both research and applications. While recent baselines struggle to generate natural and continuous facial movements due to their frame-by-frame vertex generation approach, we propose 3DFacePolicy, a pioneer work that introduces a novel definition of vertex trajectory changes across consecutive frames through the concept of ``action". By predicting action sequences for each vertex that encode frame-to-frame movements, we reformulate vertex generation approach into an action-based control paradigm. Specifically, we leverage a robotic control mechanism, diffusion policy, to predict action sequences conditioned on both audio and vertex states. Extensive experiments on VOCASET and BIWI datasets demonstrate that our approach significantly outperforms state-of-the-art methods and is particularly expert in dynamic, expressive and naturally smooth facial animations.
\end{abstract}












\maketitle

\section{Introduction}
\label{sec:intro}

Audio-driven 3D facial animation creates realistic and precise 3D facial movements with vivid and natural expressions similar to real human on 3D vertex or blendshape templates with speech input. It is widely deployed in virtual digital human, AI assistant and digital twin robot learning \cite{yang2024probabilistic}.

\begin{figure}[t]
  \centering
  \includegraphics[width=\linewidth]{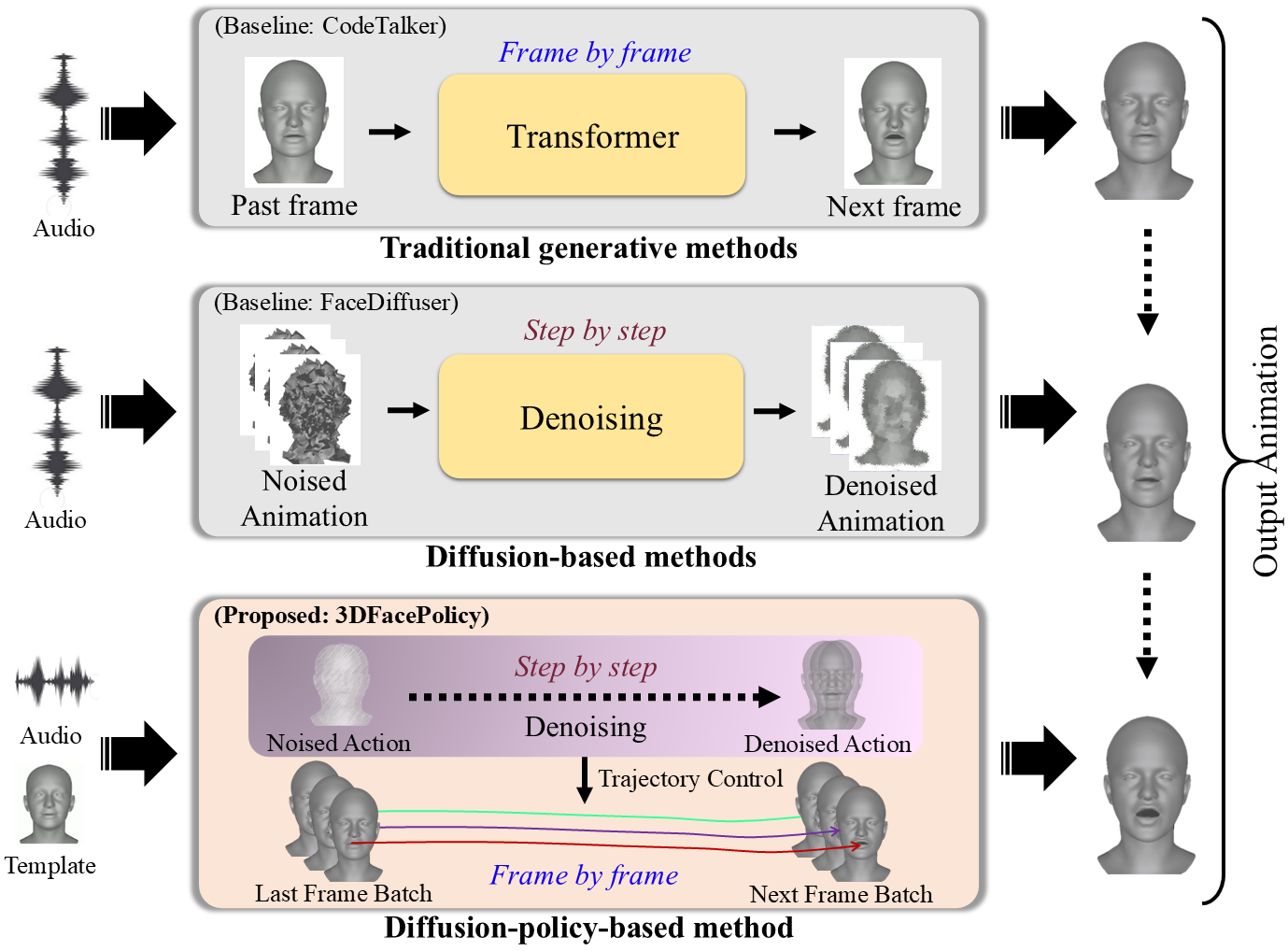}
  \caption{We propose 3DFacePolicy, a 3D facial animation architecture that controls vertex movement trajectories through diffusion policy for action prediction. Unlike traditional generative methods like CodeTalker \cite{xing2023codetalker} that predict animation frame-by-frame using Transformer structures, or diffusion-based methods like FaceDiffuser \cite{stan2023facediffuser} that generate animation from Gaussian noise step-by-step, our method reformulates vertex generation into trajectory control by accumulating denoised facial actions across consecutive frames through a robotic control mechanism.}
  \label{fig1}
\end{figure}

Recently, traditional generative methods can produce promising facial animations. Since the pioneering CNN-based approaches by \cite {Karras_Aila_Laine_Herva_Lehtinen_2017, cudeiro2019capture}, the field has evolved significantly with Transformer-based architectures \cite{fan2022faceformer, xing2023codetalker}. However, these deterministic regression methods may lead to discontinuous facial animation due to the lack of explicit restraints with masking on discrete facial regions, thus overlooking the realistic and natural human facial expressions. 

The diffusion-based methods \cite{stan2023facediffuser, sun2024diffposetalk} pioneer the integration of diffusion model to generate non-deterministic results with style conditions. However, these vertex-based generation methods may produce vague and discontinuous motions with high noise. Though 3DiFACE \cite{thambiraja20233diface} employs vertex displacements, it overlooks the smoothness modeling on vertex movement trajectories, which leads to less natural facial animations.

To address these limitations, we propose 3DFacePolicy, a novel action-based facial vertex trajectory control model that generates smooth, continuous facial animation with realistic expressions and lip-sync accuracy based on a robotic-inspired diffusion policy framework. A conceptual comparison of our method with two other mainstream approaches is shown in Fig. \ref{fig1}. We reformulate the traditional vertex generation problem as a vertex trajectory control problem by innovatively defining ``action" as temporal differential representations that encode kinematic variations between consecutive frames. This action space models both local temporal information and global spatial constraints, enabling more coherent motion synthesis compared to isolated frame-by-frame generation approaches.

For action prediction, we leverage diffusion policy \cite{chi2023diffusion,ze20243d}, a robot imitation learning framework that demonstrates high robustness on intensive and high-dimensional temporal representations. We adapt diffusion policy to predict vertex actions on 3D facial mesh, transforming facial animation synthesis from a vertex positioning problem into a motion trajectory prediction task. Through this action-based paradigm, predicted motion sequences are accumulated frame by frame to generate the final animation, naturally ensuring continuous and smooth facial motions while maintaining realistic expressions and lip-sync accuracy. 

The action sequences are first disentangled as temporal differential representations across consecutive frames, then facial movements are generated by sampling noisy action sequences from Gaussian noise and conditioning them on audio and vertex sequences using pretrained encoders, with final animation reconstructed by controlling vertex movement trajectories with the denoised action sequence. Extensive experiments demonstrate that our approach significantly outperforms state-of-the-art methods, ensuring smooth, flexible, and natural 3D facial animations.

The main contributions of our work are as follows:
\begin{itemize}
\item \textbf{A novel action-based control framework for 3D facial animation synthesis:} We design a pioneering framework that redefines 3D facial animation synthesis through an innovative action-based paradigm, introducing "action" to represent vertex trajectory changes across frames and providing a new baseline that transforms traditional vertex generation into motion trajectory control for more dynamic and natural facial motion synthesis.
\item \textbf{An insight on vertex motion smoothness through control-based prediction:} Through extensive experiments, we discover that smoother vertex motion trajectories lead to more realistic and natural facial animations. Our state-of-the-art results and trajectory visualizations validate that motion continuity achieved through control-based methods is crucial for high-quality 3D facial animation synthesis, providing valuable guidance for future research.
\item \textbf{A pioneer paradigm of treating 3D facial animation generation as facial vertex trajectory control:} We are the first to reformulate 3D facial animation synthesis as a vertex trajectory control problem by introducing robotic learning concepts. We adapt diffusion policy from robotics to treat facial dynamics as a motion control task, establishing a novel cross-domain paradigm that demonstrates the potential of applying robotic control principles to visual generation tasks.
\end{itemize}


\section{Related Work}
\label{sec:RelatedWork}

\subsection{Speech-Driven 3D Facial Animation}
Speech-driven 3D facial animation creates realistic and natural facial movements from speech, with crucial challenges in synchronizing tone, rhythm, and dynamics to mirror real-human expressions. Research has mainly focused on traditional generative methods and diffusion-based methods.

\noindent\textbf{Traditional generative methods.}
These methods design deterministic mappings from audio to facial motions using deep neural networks. Early works established foundational approaches but were limited to lip-only animations \cite{taylor2017deep,Karras_Aila_Laine_Herva_Lehtinen_2017,edwards2016jali,zhou2018visemenet}, while later research expanded to full-face animations \cite{cudeiro2019capture, richard2021meshtalk}. Transformers \cite{vaswani2017attention} then emerged as a fundamental architecture, with works \cite{fan2022faceformer, xing2023codetalker} utilizing Wav2Vec 2.0 \cite{baevski2020wav2vec} for audio processing and VQ-VAE \cite{van2017neural} inspired codebooks for motion space representation. Though these methods achieve promising results in lip synchronization and facial animation, discrete facial region processing may lead to discontinuous facial motion, and their deterministic architectures are limited in presenting dynamic facial movements \cite{stan2023facediffuser}.

\noindent\textbf{Diffusion-based methods.}
For presenting diverse facial motions, the Denoising Diffusion Probabilistic Models \cite{ho2020denoising, croitoru2023diffusion} with conditional data distribution guide are employed. In speech-driven 3D facial animation, FaceDiffuser \cite{stan2023facediffuser} is the first to integrate the diffusion model into 3D facial animation synthesis. Furthermore, works \cite{sun2024diffposetalk, thambiraja20233diface, ma2024diffspeaker} focus on the head poses and personalized styles of speakers with conditional diffusion models. These methods present diversity on facial expression \cite{MicroEmo-mm, MicroEmo-arxiv}. Nevertheless, clear facial motions and compact contextual information are overlooked, which reduces the reality and consistency of human facial motions \cite{sun2024diffposetalk}.

\subsection{Diffusion Policy Models}
The diffusion policy \cite{chi2023diffusion} is a visuomotor policy, which emerged as a crucial component in robotics for enabling agents to perform complex tasks based on visual observations such as images or depth information. Recent approaches span various paradigms including reinforcement learning \cite{levine2016end,lee2020learning}, imitation learning \cite{rajeswaran2017learning,zeng2020transporter}, and motion planning \cite{ichter2018learning,florence2021implicit,10.1007/978-3-030-27526-6_50}. Other works \cite{ke20243d, dppo2024} present multi-view condition or optimization process. 3D Diffusion Policy \cite{ze20243d} presents a two-stage architecture combining perception and decision-making, achieving state-of-the-art performance in complex manipulation tasks.

Based on these methods, our work presents a pioneering paradigm that introduces diffusion policy to facial animation synthesis. The more dynamic, natural, and continuous facial motions with vertex trajectory control are generated based on this paradigm rather than deterministic vertex positioning in traditional generative methods and blur movements in diffusion-based methods.

\section{Methodology}
\label{sec:Methods}

\begin{figure*}[t]
  \centering
  \includegraphics[width=\linewidth]{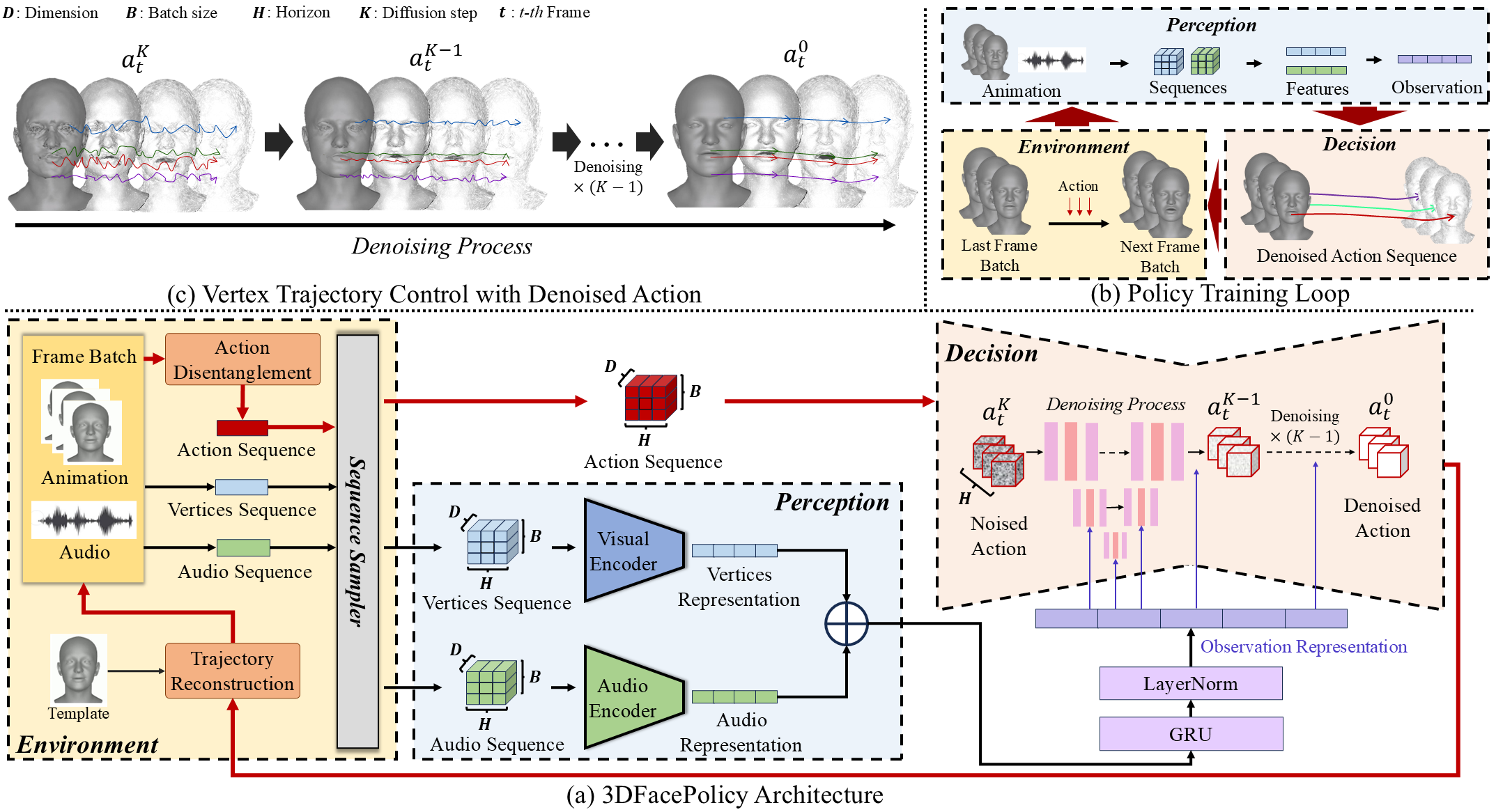}
  \caption{\textbf{Overview of 3DFacePolicy architecture.} \textit{(a)} Our architecture first disentangles animation into action sequences, then Perception module encodes vertices and audio sequences into observation representations that serve as conditions for Decision module, where actions are produced through denoising and control vertex trajectories on template to output animation.  \textit{(b)} The Environment, Perception, and Decision form a policy training loop for vertex trajectory control. \textit{(c)} Noised action sequence sampled from Gaussian noise is gradually denoised into smooth and ordered actions for vertex trajectory control.}
  \label{method}
\end{figure*}

\subsection{Problem Formulation}
In our method, we design facial movements diffusion policy model (3DFacePolicy) to control the trajectory of vertex movements in consecutive frames, represented as the action $a_{0}^{1:N} = (a_{0}^{1},a_{0}^{2},...,a_{0}^{N})\in \mathbb{R}^{N\times V\times 3}$, where $N, V, 3$ are the number of frames, mesh vertices, and dimensions. Conditioning on audio $s^{1:N}$ and vertices states $x^{1:N}$, the action $a_{t}^{1:N}$ from Gaussian Noise is gradually denoised into noise-free action sequence $a_{0}^{1:N}$, where $t\in{\{1,...,T\}}$ is the diffusion step. Therefore, our goal of the proposed architecture 3DFacePolicy is to control the movement trajectory of vertices with denoised action based on the conditional input of audio and vertices state. The problem could be formulated as:
\begin{equation}
    a_{0} = 3DFacePolicy(a_{t},s,x,t),
\end{equation}
With the predicted action sequence $a_{0}$ and the vertices of mesh template $x_{temp}$, the vertices in $n$-$th$ frame is presented as:
\begin{equation}
    x_{0}^{n} = {x}_{temp} + \sum_{i=1}^{n}a_{0}^{i}, n\in{\{1:N\}},
    \label{output_ani}
\end{equation}
where ${x_{0}^{n}}$ is the $n$-$th$ frame of output animation $x_{0}^{1:N}$ with audio input following frame-by-frame predicted actions.


\subsection{Architecture}

\subsubsection{Overview.}
We design our model with three modules following the policy training loop in robotics: Environment, Perception, and Decision, as illustrated in Fig. \ref{method} (b). Environment module controls vertex trajectories with predicted action sequences and disentangles action sequences for next frame batches. Perception module generates comprehensive observation representation from vertices and audio sequences, serving as conditional input for guiding action policy learning process in Decision module. Decision module presents a conditioned denoising process (Fig. \ref{method} (c)) where random and disordered action $a^K_t$ are gradually denoised into temporally and spatially ordered action sequence $a^0_t$ conditioning on observations, ensuring smooth and natural motion trajectories for every vertex to generate dynamic and realistic facial animations. The overall architecture is shown in Fig. \ref{method} (a).

\subsubsection{Environment.}
The Environment module disentangles the action sequence from animation and samples the vertices, action, and audio sequence into a limited duration, ensuring the policy is trained in an action space with intensive context to maintain motion consistency and accuracy. 

\textit{Action:} A critical component of our approach is the formulation of action sequences that effectively capture facial motion dynamics. We define actions as temporal differential representations that encode the kinematic variations with adaptive scaling mechanisms between consecutive frames. This motion-centric formulation enables our model to learn smooth vertex movement trajectories.

Given a facial animation sequence with vertices $x^{1:N} = \{x^1, x^2, ..., x^N\} \in \mathbb{R}^{N \times V \times 3}$, the fundamental temporal displacement operator is expressed as:
\begin{equation}
    \mathcal{D}_{temporal}^{n}  = x^{n+1} - x^{n}, n\in[0,N-1]
\end{equation}
This operator captures the raw inter-frame motion vectors that serve as the basis for action formulation.

Then we apply an adaptive scaling mechanism to enhance motion sensitivity and stability. An exponential weighting factor is leveraged to adaptively modulate the scaling based on the motion intensity. The scaling factor is defined as:
\begin{equation}
    \Lambda_{adaptive}  = \exp(-\beta\cdot\| \mathcal{D} _{temporal}^{n} \| _{F}^{2}) 
\end{equation}
The exponential function provides natural motion-based weighting where $\beta$ controls the sensitivity decay rate, ensuring that subtle facial movements receive higher scaling weights while preventing large motions. $\| \cdot \| _{F}$ denotes the Frobenius norm. Finally, the action is formulated as follows.
\begin{equation}
    a^{n}  = \varepsilon_{scaling}\cdot \Lambda_{adaptive} \cdot \mathcal{D}_{temporal}^{n}
\end{equation}
where $\varepsilon_{scaling}$ is the base scaling parameter. The differential representation naturally preserves motion continuity through accumulation. Through empirical evaluation, this action-formulation-based method outperforms traditional vertex generation approaches.

Then after frame length alignment, vertices sequence $x^{1:N}$, audio sequence $s^{1:N}$, and action sequence $a^{1:N}$ with the same sequence length $N$ are generated. Then sequence sampler samples data into manageable fixed durations called Horizon $H$, ensuring the action prediction policy is trained in a relatively local context with observation condition length $N_{obs}$ and action-making length $N_{act}$ to maintain smoothness and consistency. The sampled action sequence is isolated for coordination through the diffusion policy in the Decision module, while others serve as observation representations in the Perception module.

\subsubsection{Perception.}
Perception module transforms the vertices and audio sequences with $H$ length into a comprehensive representation $O=\{O_{x},O_{s}\}$ that serves as conditions for Decision module, considered as a temporal observation fraction containing both audio and visual features. For visual data, we employ a lightweight encoder architecture inspired by \cite{stan2023facediffuser}, comprising linear layer, convolutional layer, and max-pooling layer to downsample the 3D features into a 1024-dimensional representation. The audio encoder utilizes the pretrained HuBERT model \cite{hsu2021hubert} to generate audio representations. The visual and audio features are concatenated and processed through a GRU layer followed by LayerNorm to generate comprehensive observation representations for Decision module. Maintaining spatial coherence across facial regions draws on principles from \cite{Panoptic-wacv, Panoptic-tcsvt}.

\subsubsection{Decision.}
The conditional denoising diffusion model is the backbone for learning facial action policy following \cite{ze20243d}. For $K$ iterations, A noised action sequence $a_{K}$ is sampled from Gaussian noise with $H$ length, conditioning on visual features $x$ and audio features $s$, it is gradually denoised into a smooth and ordered action sequence $a_{0}$ with reverse process. The equation is formulated as follows:
\begin{equation}
    a_{k-1}=\alpha_{k}(a_{k}-\gamma_{k}\epsilon_{\theta}(a_{k},k,x,s))+\sigma_{k}\mathcal{N}(0,\mathrm{I}),
\end{equation}
where $\epsilon_{\theta}$ is the denoising network, $\alpha_{k}$, $\gamma_{k}$ and $\sigma_{k}$ are functions of $k$ iteration. $\mathcal{N}(0,\mathrm{I})$ is Gaussian noise. After $K$ iterations, the denoised action sequence is predicted.

For vertex trajectory reconstruction, with predicted action sequence ${\mathcal{a}}^{1:N}_0$, the final vertex positions are computed as:
\begin{equation}
    x_{0}^{n} = {x}_{temp} + \sum_{i=1}^{n}\mathcal{G}({a}^{i}_0)
\end{equation}
where ${x}_{temp}$ is the neutral template mesh, $\mathcal{G}(\cdot)$ is the inverse transformation function, which presented as:
\begin{equation}
    \mathcal{G}(a^{i}) = a^{i} / (\varepsilon_{scaling} \cdot \Lambda_{adaptive})
\end{equation}

\subsubsection{Loss Function.}
In the diffusion process, the noise $\epsilon^{\theta}$ is added on a randomly sampled action sequence $a_{0}$ at $k$ iteration to train the denoising network $\epsilon_{\theta}$. The objective of this process is to predict the noise added on the sequence, which is presented as diffusion loss in our model:
\begin{equation}
    \mathcal{L}_{\rm{diff}}=\mathrm{MSE}(\epsilon^{k},\epsilon_{\theta}(\overline{\alpha_{k}}a_{0}+\overline{\beta_{k}}\epsilon^{k},k,x,s)),
\end{equation}
where $\overline{\alpha_{k}}$ and $\overline{\beta_{k}}$ are noise schedules during diffusion steps. However, diffusion-loss-only training is defective for generating vertices sequences with smooth actions. Here We also use reconstruction loss on vertex space to supervise the visual output:
\begin{equation}
    \mathcal{L} _{\rm{rec}} = \mathbb{E}_{n}[\frac{1}{N}\sum_{n=1}^{N}{\| x_{0}^{n} - \hat{x}_{0}^{n} \|}^{2}],
\end{equation}
where $x_{0}^{n}$ and $\hat{x}_{0}^{n}$ are the predicted vertices sequence $x_{0}$ and ground truth $\hat{x}_{0}$ in the $n$-$th$ frame from frame length $N$. We also use velocity loss to enhance the action smoothness:
\begin{equation}
    \mathcal{L} _{\rm{vel}} = \mathbb{E}_{n}[ \frac{1}{N}\sum_{n=1}^{N}{\| (x_{0}^{n-1}-x_{0}^{n})  - (\hat{x}_{0}^{n-1}-\hat{x}_{0}^{n}) \|}^{2} ].
\end{equation}
The total loss is the sum of these three losses:
\begin{equation}
    \mathcal{L} = \lambda_{1}\mathcal{L}_{\rm{diff}} + \lambda_{2}\mathcal{L} _{\rm{rec}} + \lambda_{3}\mathcal{L} _{\rm{vel}}.
\end{equation}
Here, $\lambda_{1}$, $\lambda_{2}$, and $\lambda_{3}$ are weight coefficients that balance the relative importance of diffusion loss, reconstruction loss, and velocity loss in the total loss function.

\subsubsection{Implementation details.} We use DDIM \cite{song2020denoising} as denoising scheduler and sample prediction. For the sequence sampler, we set horizon length $H = 16$, $N_{obs} = N_{act} = 8$ and $\beta = 0.1$ in action scaling. For the trade-off parameters in loss function, $\lambda_{1}$, $\lambda_{2}$, $\lambda_{3}$ are empirically set to $1$, $2$, $0.5$ respectively. We train 600 epochs on 32 batch size. The observation representation is 1024 dimensions. Our model is trained on a single V100 GPU with 32GB RAM. We employed the AdamW optimization algorithm, setting the learning rate to 0.0001 and gradually decreased to 0.000001.

\begin{table*}[t]
\centering
\caption{Quantitative comparison of 3DFacePolicy with state-of-the-art methods on VOCASET and BIWI datasets. 
}
\label{Quan}
\begin{tabular}{l@{\hspace{2.5mm}}ccclcc@{\hspace{2.5mm}}c}
\toprule
\multicolumn{4}{c}{VOCASET} & \multicolumn{4}{c}{BIWI} \\
\cmidrule(lr){1-4} \cmidrule(lr){5-8}
\multirow{2}{*}{Method} & MVE $\downarrow$ & FDD $\downarrow$ & UFVE $\downarrow$ & \multirow{2}{*}{Method} & MVE $\downarrow$ & FDD $\downarrow$ & UFVE $\downarrow$ \\
& ($10^{-3}mm$) & ($10^{-7}mm$) & ($10^{-3}mm$) & & ($10^{-3}mm$) & ($10^{-5}mm$) & ($10^{-3}mm$) \\
\midrule
VOCA & 0.983 & 2.662 & - & VOCA & 8.361 & 7.532 & - \\
FaceFormer & 0.935 & 2.163 & 0.497 & FaceFormer & 7.275 & 4.006 & 6.908 \\
CodeTalker & 0.888 & 2.258 & 0.471 & CodeTalker & 7.378 & 4.215 & 7.005 \\
FaceDiffuser & 0.901 & 2.437 & 0.477 & FaceDiffuser & 6.809 & 3.910 & 6.543 \\
ScanTalk & 0.861 & 2.101 & - & UniTalker & \textbf{6.417} & 5.044 & 6.148\\
3DFacePolicy & \textbf{0.847} & \textbf{1.502} & \textbf{0.416} & 3DFacePolicy & 7.167 & \textbf{1.778} & \textbf{5.433} \\
\bottomrule
\end{tabular}
\end{table*}

\section{Experiments}
\label{sec:Experiment}

We conduct comprehensive experiments to evaluate 3DFacePolicy on VOCASET \cite{cudeiro2019capture} and BIWI \cite{Fanelli_Gall_Romsdorfer_Weise_Gool_2010}. Our evaluation includes both qualitative and quantitative analysis. We also conduct vertex trajectory smoothness comparison to further evaluate the contribution of our method. User study is also conducted as a convincing measurement to evaluate our method based on audiovisual user experience. Several ablation studies are also designed to analyze the impact of key settings in our model, specifically examining the effects of action definition, diffusion policy, horizon length, and loss function choices. Additional experimental results, model computation efficiency analysis, inference time, and video demonstrations are provided in the \textbf{Supplementary Material}.

\subsection{Experimental Settings}

\subsubsection{Baseline.}
We compare 3DFacePolicy with state-of-the-art methods: VOCA \cite{cudeiro2019capture}, FaceFormer \cite{fan2022faceformer}, CodeTalker \cite{xing2023codetalker}, FaceDiffuser \cite{stan2023facediffuser}, ScanTalk \cite{nocentini2024scantalk3dtalkingheads}, UniTalker \cite{fan2024unitalker} and KmTalk \cite{xu2024kmtalk}. For fair comparison, we use their official implementations and follow the same training/testing splits of datasets. All methods are evaluated under identical experimental settings to ensure valid comparisons.

\subsubsection{Dataset.}
\textit{VOCASET dataset.}~\cite{cudeiro2019capture} Contains 480 3D facial animation sequences with facial motions and audio from 12 subjects, recorded at 60 frames per second with 3-4 seconds duration. The template mesh uses FLAME \cite{li2017learning} topology with 5023 vertices. We use the same training set (VOCA-Train), validation set (VOCA-Val) and test set (VOCA-Test) as in \cite{xing2023codetalker, fan2022faceformer}.
\textit{BIWI dataset.}~\cite{Fanelli_Gall_Romsdorfer_Weise_Gool_2010} Contains recordings of 40 English sentences from 14 subjects, each read twice (neutral and emotional), captured at 25 frames per second with average duration of 4.67 seconds. The 3D mesh template contains 23370 vertices. We use only the emotional sequences and follow the same dataset split as in \cite{ma2024diffspeaker, fan2022faceformer}.

\subsubsection{Evaluation Metric.}
To comprehensively evaluate the quality of generated facial animations, we employ Mean Vertex Error (MVE), Facial Dynamics Deviation (FDD) and Upper-face Vertex Error (UFVE) as our evaluation metrics following \cite{fan2024unitalker}. the MVE and UFVE measures the deviation of all face vertices and upper-face vertices respectively, while FDD measures the variation of facial dynamics for a motion sequence in comparison with ground truth. The calculation formulas of three metrics are listed in the supplementary material.


\begin{figure*}[t]
  \centering
  \includegraphics[width=\linewidth]{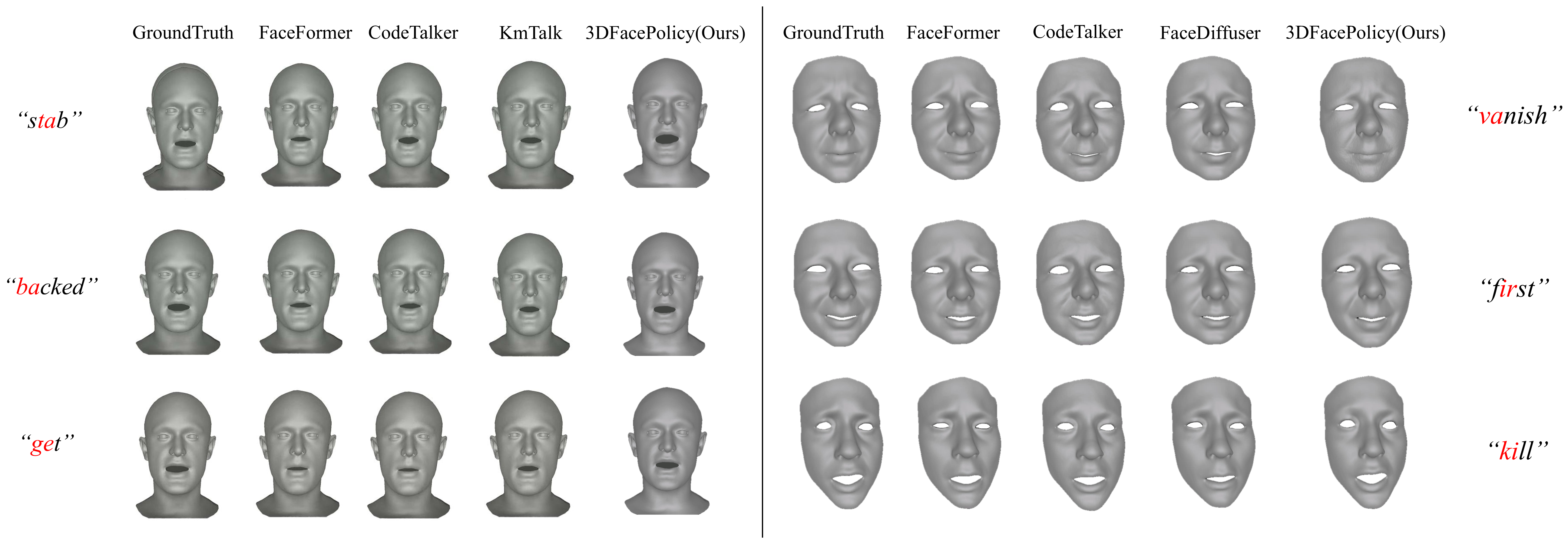}
  \caption{Qualitative comparison of facial animation results on VOCASET (left) and BIWI (right). The figure shows the results of six different phonemes compared with other state-of-the-art results.}
  \label{Qualitative}
\end{figure*}

\subsection{Quantitative Evaluation}
\label{secsec:quan_eva}

We conduct comprehensive quantitative evaluations on VOCASET and BIWI datasets, comparing our method with state-of-the-art approaches. Table \ref{Quan} presents the comparison results using Mean Vertex Error (MVE), Facial Dynamics Deviation (FDD) and Upper-face Vertex Error (UFVE) metrics.

For the VOCASET dataset, our method achieves the best performance in all metrics with a notable reduction in FDD. For the BIWI dataset, we achieve the best FDD and UFVE score, though with a slightly higher MVE score. This trade-off demonstrates our method's emphasis on capturing dynamic facial movements over strict vertex positioning accuracy. These results demonstrate 3DFacePolicy's effectiveness in modeling facial dynamics with motion trajectory control across datasets, validating our goal of producing more smooth and expressive facial animations while preserving temporal consistency and motion naturality.

\begin{table*}[t]
  \centering
  \caption{User study results. Preference percentage of A/B testing on Lip Sync, Realism and Emotional Expression is shown.}
  \label{user}
  \begin{tabular}{lcc|cc|cc}
    \toprule
    \multirow{2}{*}{Methods} & \multicolumn{2}{c}{Lip Sync (\%) $\uparrow$} & \multicolumn{2}{c}{Realism (\%) $\uparrow$} & \multicolumn{2}{c}{Emotional Expression (\%) $\uparrow$} \\
    & Ours & Competitor & Ours & Competitor & Ours & Competitor \\
    \midrule
    Ours \textit{vs.} CodeTalker & \textbf{53.85} & 46.15 & \textbf{56.92} & 43.08 & \textbf{67.69} & 32.31 \\
    Ours \textit{vs.} FaceDiffuser & \textbf{63.08} & 36.92 & \textbf{78.46} & 21.54 & \textbf{84.62} & 15.38 \\
    Ours \textit{vs.} GT & 41.54 & \textbf{58.46} & \textbf{52.31} & 47.69 & \textbf{52.31} & 47.69 \\
    \bottomrule
  \end{tabular}
\end{table*}
\subsection{Qualitative Evaluation}

We visually evaluate our method against state-of-the-art methods including FaceFormer \cite{fan2022faceformer}, CodeTalker \cite{xing2023codetalker}, FaceDiffuser \cite{stan2023facediffuser} and KmTalk \cite{xu2024kmtalk}. The results are rendered on FLAME template of VOCASET and BIWI shown in Fig. \ref{Qualitative}. 

Our method demonstrates natural and expressive facial movements with more realistic lip shapes during speech. When pronouncing syllables like ``ki" in ``kill" and ``ba" in ``backed", our method produces more realistic results than other methods, with mouth movements closely matching ground truth. Moreover, for expressive syllables like ``va" in ``vanish" and ``ir" in ``first", our method shows clearer emotions while maintaining natural mouth shapes than other baselines. The diverse facial movements and variable expressions are more explicit than existing methods, indicating that our approach prioritizes synthesizing dynamic expressions and realistic facial movements with vertex trajectory control over strict positioning.

\subsection{Vertex Motion Smoothness Evaluation}
We conduct experiments to visually evaluate the smoothness of facial vertex trajectories on generated animations over time with other state-of-the-art methods on BIWI dataset, randomly selecting three vertices in the lip region where motion amplitude is typically large and visualizing their movement trajectories. \textbf{Note:} video demonstrations of vertex trajectories are provided in supplementary materials. 

By observing facial motion trajectories during high-amplitude segments such as the ``re" sound in ``read", our method produces significantly smoother transitions while other baselines exhibit uneven variations and noticeable abrupt shifts. These less smooth trajectories from vertex generation-based methods lead to subtle jittering in facial animations, with accumulation of such artifacts resulting in less natural and realistic outcomes. Our trajectory control-based approach effectively addresses these issues by controlling facial vertex movements that better approximate real human facial motion, demonstrating that smoother vertex trajectories directly contribute to more natural facial animation quality.

\begin{figure*}[t]
  \centering
  \includegraphics[width=\linewidth]{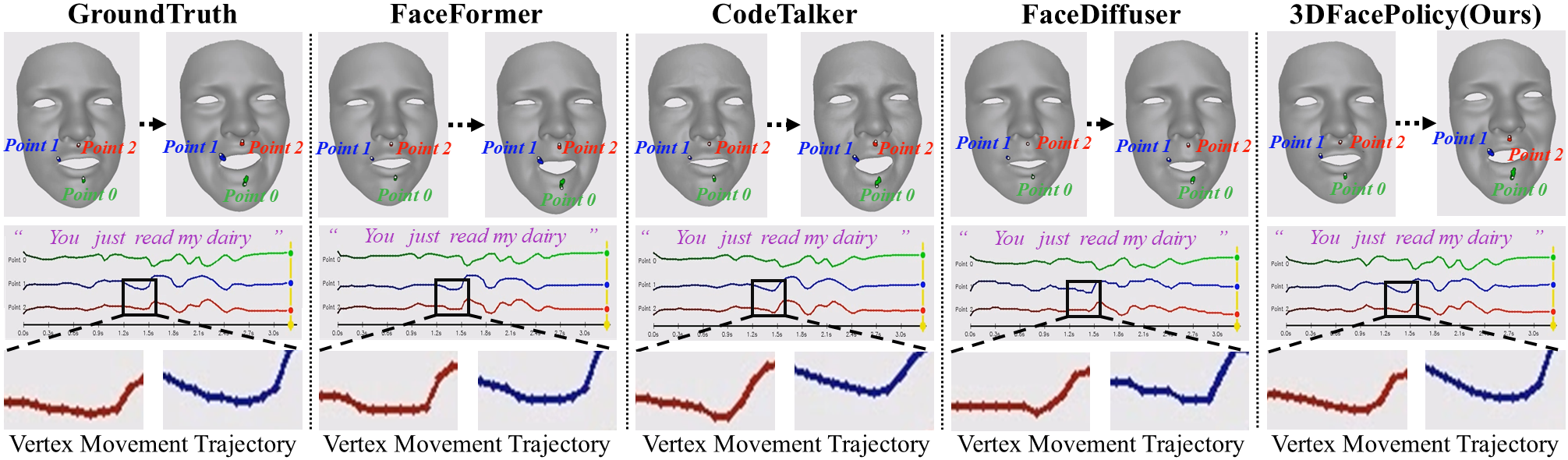}
  \caption{Vertex movement trajectory comparison on BIWI dataset. Three randomly chosen vertex trajectories with high motion amplitude in lip region are printed to compare the animation smoothness of 3DFacePolicy with other baselines.}
  \label{smooth}
\end{figure*}


\subsection{User Study}
We conduct a user study to evaluate the quality of generated 3D talking faces following \cite{fan2024unitalker}, selecting CodeTalker \cite{xing2023codetalker}, FaceDiffuser \cite{stan2023facediffuser}, and ground truth as competitors in an A/B test. Participants watch randomly arranged paired videos of 3DFacePolicy and other works, evaluating three key metrics: lip synchronization, realism, and emotional expression. We randomly sampled 30 examples from VOCASET and BIWI with 15 video pairs per participant, resulting in 450 effective evaluation entries from 30 participants with good visual and auditory quality. Table \ref{user} shows the statistical comparison of user preferences. Our method is preferred in all three metrics, greatly outperforming other state-of-the-art methods in realism and emotional expression while maintaining competitive performance with ground truth. For lip synchronization, 3DFacePolicy shows better preference than other baselines and slightly worse than ground truth as expected. Overall, 3DFacePolicy demonstrates explicit and realistic facial expressions, proving more natural and suitable for 3D facial animation through our facial motion control approach.


\subsection{Ablation Study}

We evaluate the effect of key elements in our proposed model: \textit{(i) Action Definition.} Evaluating the impact of our key contribution on 3D facial animation synthesis; \textit{(ii) Diffusion Policy.} Evaluating the robotic controlling mechanism; \textit{(iii) Horizon Length.} Evaluating the impact of horizon length choice; \textit{(iv) Loss Function.} Evaluating individual contributions of loss function components to model performance. Each experiment trains on VOCASET with MVE, FDD and UFVE units of $(\times{10^{-3}}mm)$, $(\times{10^{-7}}mm)$ and $(\times{10^{-3}}mm)$ respectively. Results are shown in Table \ref{tab:ablation}.

\subsubsection{Verification for Action definition choice.}
To verify the effectiveness of our proposed vertex action definition, we perform an ablation study by removing the scaling factor and whole action disentanglement module, which predicts actions without adaptive scaling and directly predicts facial animation in vertex space without action sequences.

Results show that removing action disentanglement significantly degrades performance across all metrics, highlighting its critical role in our framework. By separating actions from animation, the model can effectively control vertex trajectories conditioned on both audio and vertex states with smooth action sequences. Without action, the model struggles to establish clear relationships between audio input and facial movements, resulting in less accurate and natural facial animations, confirming that action is essential for generating accurate, smooth, and expressive facial animations and validates the effectiveness of reformulating generation into a controlling problem in 3D facial animation.

\begin{table}[t]
\centering
\caption{Ablation studies on action definition choice, diffusion policy, horizon length, and loss function components.}
\label{tab:ablation}
\begin{tabular}{lccc}
\toprule
\textbf{Action Definition Choice} & MVE $\downarrow$ & FDD $\downarrow$ & UFVE $\downarrow$ \\
\midrule
w/o Action & 5.278 & 11.977 & 4.913 \\
w/o Adaptive Scaling & 0.971 & 2.582 & 0.568 \\
\midrule
\textbf{Diffusion Policy} & & & \\
\midrule
w/o Diffusion Policy & 1.344 & 5.771 & 0.908 \\
\midrule
\textbf{Horizon, Observation,} & & & \\
\textbf{Action Steps} & & & \\
\midrule
8, 4, 4  & 1.207 & 4.260 & 0.653 \\
24, 12, 12 & 0.972 & 2.504 & 0.637 \\
\midrule
\textbf{Loss Function} & & & \\
\midrule
w/o $\mathcal{L}_{\rm{rec}}$ & 1.074 & 1.786 & 0.885\\
w/o $\mathcal{L}_{\rm{vel}}$ & 0.887 & 1.705 & 0.434\\
w/o $\mathcal{L}_{\rm{diff}}$ & 0.930 & 2.587 & 0.604\\
\midrule
Full Model & \textbf{0.847} & \textbf{1.502} & \textbf{0.416}\\
\bottomrule
\end{tabular}
\end{table}

\subsubsection{Verification for Diffusion Policy.}
To evaluate the effect of diffusion policy in 3D facial animation synthesis, we design a plain diffusion method without policy component that directly predicts entire action sequences with full animation length using only diffusion model, isolating the policy loop component's contribution and compare with other diffusion-based methods at action space level. 

Results show that our diffusion policy-based method comprehensively outperforms the plain diffusion model. It demonstrates that directly inferring entire action sequences without the policy component lacks intensive contextual information, leading to unstable vertex trajectory control. It confirms that the robotic control methodology provides a stable and efficient paradigm for visual generation tasks such as 3D facial animation by transforming the problem from vertex generation to trajectory control, better addressing the smoothness of generated vertex trajectories and naturalness of facial motions.

\subsubsection{Verification for Horizon Length.}
The horizon length determines the temporal context window for action prediction in facial motion synthesis. We evaluated three different settings of horizon $H$, observation condition length $N_{obs}$ and action-making length $N_{act}$, while $N_{obs}$ equals $N_{act}$ and is half the length of horizon $H$.

Horizon of 8 frames leads to less accurate predictions due to insufficient temporal context, while a longer horizon (24 frames) shows insufficient performance by overlooking intensive context information. The horizon length of 16 frames achieves optimal performance, particularly in capturing dynamic facial movements. Therefore, we set $H = 16$ as our default configuration.

\subsubsection{Verification for Loss Function.}
In this experiment, we remove reconstruction loss $\mathcal{L}_{\rm{rec}}$, velocity loss $\mathcal{L}_{\rm{vel}}$ and diffusion loss $\mathcal{L}_{\rm{diff}}$ respectively to assess the impact of each loss.

Without $\mathcal{L}_{\rm{rec}}$, both MVE and UFVE show surprising increases, indicating its role in maintaining vertex accuracy. Removing $\mathcal{L}_{\rm{vel}}$ primarily affects FDD, suggesting its importance for temporal consistency. The absence of $\mathcal{L}_{\rm{diff}}$ leads to a substantial increase in FDD and also effects MVE and UFVE, demonstrating its crucial part in maintaining smooth facial motions and model stabilization. Each component contributes meaningfully to the model's ability to predict dynamic, natural, and accurate facial animations.


\section{Conclusion}
\label{sec:Conclusion}

In this paper, we propose 3DFacePolicy, a pioneering approach that reformulates Audio-driven 3D facial animation from vertex generation to action-based trajectory control by introducing ``action" as temporal differential representations encoding frame-to-frame vertex movements. We are the first to adapt diffusion policy from robotics to predict smooth action sequences conditioned on audio and vertex states, naturally ensuring smooth and realistic facial motions with accurate lip synchronization. Comprehensive experiments on VOCASET and BIWI datasets demonstrate that 3DFacePolicy significantly outperforms state-of-the-art approaches, particularly validating our insight that smoother vertex movement trajectories directly contribute to more natural animations. Future work will focus on adaptive sequence sampling strategies to enhance model flexibility while maintaining motion naturalness.


\begin{figure*}[t]
  \centering
  \includegraphics[width=\linewidth]{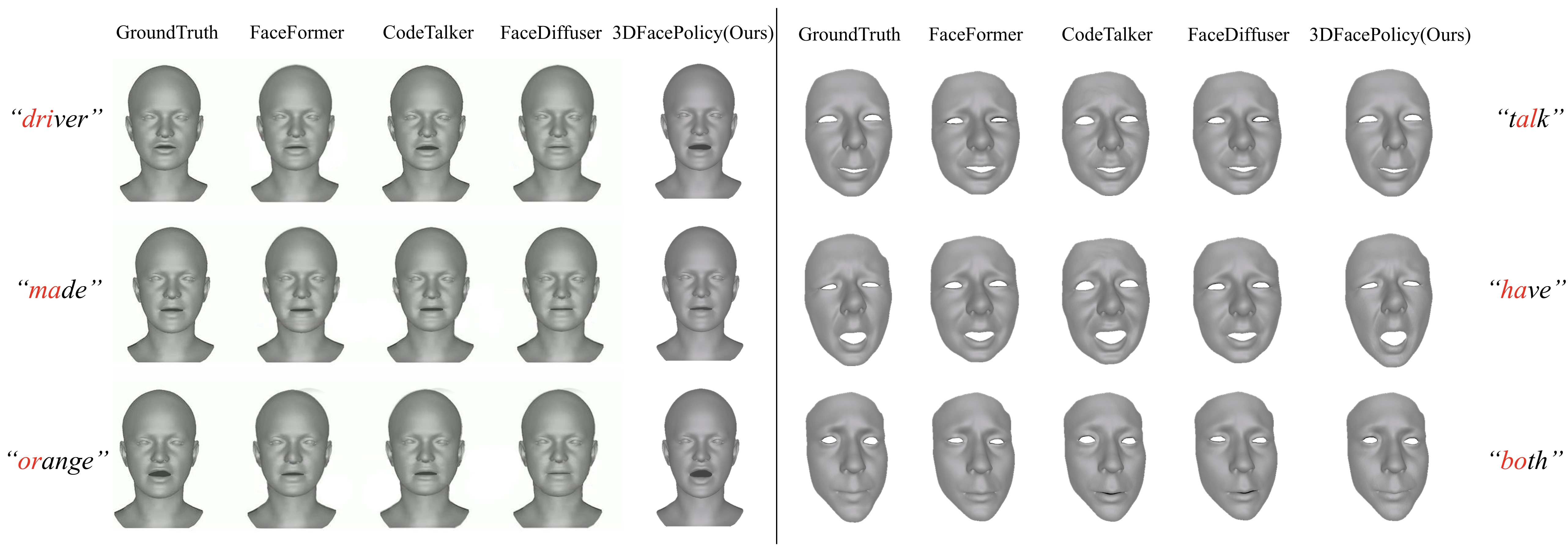}
    \caption{Additional experiment results on both VOCASET (left) and BIWI (right) datasets. For VOCASET results, our method generates more accurate lip shapes and natural facial movements compared to other approaches on challenging syllables. For BIWI results, our method better captures the emotional expressions of speech, producing more expressive facial animations while maintaining natural facial movements.}
    \label{additional}
\end{figure*}

\section{Supplementary Material}
\subsection{Overview}
In this supplementary, we first provide additional experimental results of VOCASET and BIWI with the comparison between our approach and other state-of-the-art methods in Sec. Additional Experimental Results. In Sec. Evaluation Metrics, we list the calculation formula of metrics. In Sec. Model Efficiency, we discuss the model efficiency of our method compared with other baselines.
In Sec. Model Reproducibility, we provide the detailed network architecture of our Observation Encoder and Diffusion Module for model reproducibility.
In Sec. Failure Cases, The failure cases of our proposed method are discussed.
Finally, we present the details of the user study in Sec. User Study.
Note that we provide our code and video comparison results attached together with supplementary material. Please check more details of our \textbf{video demonstration results in the attached file}. 

\subsection{Additional Experimental Results}
\label{sec:add}

Due to the limited space in Sec. Experiments, we present additional qualitative comparison results from both the VOCASET and BIWI datasets in Fig. \ref{additional}. These supplementary results further demonstrate our method's superior performance in generating natural facial movements and emotional expressions compared to existing approaches.

The results on VOCASET show that our method's enhanced capability in capturing detailed facial movements and dynamic lip shapes. Our model shows significant improvement in reproducing challenging syllables shown in the accurate lip shapes for the \textit{``ma"} and \textit{``or"} sounds in \textit{``made"} and \textit{``orange"} respectively. Our method also excels in generating large lip movements for emphasized sounds, as demonstrated in the \textit{``dri"} syllable in \textit{``driver"}. These results verify 3DFacePolicy's ability to generate more dynamic and natural facial animations compared to current state-of-the-art methods.

The results from BIWI dataset further highlight our method's effectiveness in emotional expression generation. Our approach shows superior performance in capturing and reproducing emotional expressions, as evidenced by the appropriate sorrow expression when pronouncing \textit{``have"} and the delightful expression in \textit{``talk"}. These results demonstrate 3DFacePolicy's strength in generating accurate facial expressions that match the emotional content of audio input, showing its ability to handle both facial movement and emotional expression in facial animation synthesis.

\subsection{Evaluation Metrics}
\label{sec:eva}
More detailed metric calculation description is presented as follows:

\textit{i) Mean Vertex Error (MVE):} This metric evaluates the overall geometric accuracy by calculating the average L2 distance between predicted vertices and ground truth vertices:
\begin{equation}
    \text{MVE} = \frac{1}{NV}\sum_{n=1}^{N}\sum_{v=1}^{V}\|x_{n,v} - \hat{x}_{n,v}\|_2,
\end{equation}
where $V$ is the number of vertices in the mesh, and $N$ is frame number. $x_{n,v}$ and $\hat{x}_{n,v}$ represent the predicted and ground truth vertex $v$ at frame $n$ respectively.

\textit{ii) Facial Dynamics Deviation (FDD):} This metric calculates the deviation between the predicted and ground truth upper-face dynamics relative to the template mesh:
\begin{equation}
    \text{FDD} = \frac{1}{N}\sum_{n=1}^{N}\|(x_n - x_{temp}) - (\hat{x}_n - x_{temp})\|_2,
\end{equation}
where $x_n$ and $\hat{x}_n$ represent the predicted vertex and ground truth vertex at frame $n$. $x_{\text{temp}}$ represents the template mesh vertices. This metric specifically evaluates how well the model captures the dynamic movements of the upper face region relative to the neutral template pose.

\textit{iii) Upper-face Vertex Error (UFVE):} This metric evaluates the vertex deviation of upper face by calculating the average L2 distance between predicted vertices and ground truth vertices:
\begin{equation}
    \text{UFVE} = \frac{1}{N{V_u}}\sum_{n=1}^{N}\sum_{v=1}^{V_u}\|x_{n,v} - \hat{x}_{n,v}\|_2,
\end{equation}
where $V_u$ is the number of vertices in the upper face mesh.

\subsection{Model Efficiency}
\label{sec:eff}
The model efficiency will influence practical applications, it is mainly measured from computational cost, model complexity and processing time.
For the computational cost, we used floating-point operations (FLOPs) as the evaluation indicators. For the processing time, we calculate the average processing times (Avg PT) per frame for different models of competing baselines.

As shown in Table \ref{efficiency}, we present the model efficiency comparisons between our scores and the best baselines' scores, here we list the competing transformer-based baseline CodeTalker \cite{xing2023codetalker} and diffusion-based baseline FaceDiffuser \cite{stan2023facediffuser}.
From the quantitative result of FLOPs, 3DFacePolicy demonstrates substantially reduced computational complexity comparing to other two baselines, as our method directly predicts vertex trajectories in action space rather than computing frame-by-frame vertex positions through complex transformer operations. This more efficient representation and processing strategy leads to lower computational requirements. For the average processing time, our model achieves the fastest animation prediction time while maintaining superior generation quality, demonstrating the effectiveness of our diffusion policy strategy and action-based framework. Overall, these efficiency metrics indicate that our method achieves better performance with superior computational cost and processing speed. For real-time synthesizing, current 3D animation generation works consistently relies on offline video supervised learning. Though our algorithm achieves superiority, realization of real-time inference of this field might still need computational resources breakthrough in short time. 

\begin{table}
\begin{center}
{
\caption{The floating-point operations (FLOPs) evaluate the computational cost; The average processing time (Avg PT) per frame evaluates processing speed. The lower the better.}
\label{efficiency}
\begin{tabular}{ccc}
\toprule
\multirow{2}{*}{Model} & FLOPs & Avg PT \\
& \thinspace(G) & \thinspace(ms) \\
\midrule
CodeTalker & $40.37$ & $12.49$  \\
FaceDiffuser & $31.20$ & $169.30$ \\
Ours & $\bf{19.80}$ & $\bf{11.60}$ \\
\bottomrule
\end{tabular}
}
\end{center}
\end{table}

\subsection{Model Reproducibility}
\label{sec:repro}
For the reproducibility of our model, here we detail our Perception and Decision module as follows.

The Perception module represents the vertices sequence and audio sequence into observation conditions. It includes visual encoder and audio encoder. The visual encoder architecture is inspired by \cite{stan2023facediffuser}, using a series of downsampling operations through linear layers, convolutions and max pooling to process the 3D vertex features. The visual processing module also benefits from prior experience in \cite{uneven}. The audio encoder leverages the pretrained \textit{hubert-large-ls960-ft} version of HuBERT \cite{hsu2021hubert}, which has shown superior performance compared to Wav2Vec 2.0 \cite{Baevski_Zhou_Mohamed_Auli_2020} for speech processing tasks \cite{haque2023facexhubert}. The HuBERT model employs a temporal convolutional feature extractor followed by a multi-layer transformer encoder to generate robust audio representations. The concept of cross-modal processing is partly inspired by previous work on \cite{Thermal-to-Color}.

For feature concatenation, we choose gated recurrent unit (GRU) over Long short-term memory (LSTM) due to its more efficient handling of temporal dependencies in sequential data. The GRU layer processes the concatenated visual and audio features. LayerNorm helps stabilize training by normalizing the hidden states, which is particularly important to provide different scales of multimodal features. This architecture ensures effective integration of spatial and temporal information while maintaining computational efficiency.

The Decision module predicts actions of vertices from Gaussian noise with diffusion denoising process. The design of this process follows \cite{ze20243d}, which adopts a 1D temporal CNN architecture based on a U-Net backbone. It consists of three progressive down-sampling blocks with channel dimensions expanding from 256 to 1024, followed by symmetric up-sampling blocks with skip connections. We employ Feature-wise Linear Modulation (FiLM) conditioning mechanism at down, mid, and up levels for comprehensive feature modulation, where the global conditioning integrates both audio and visual context through diffusion step embedding. The temporal convolutions use kernel size of 3 and 8 group normalization, which provides an optimal balance between capturing temporal dependencies and maintaining computational efficiency.

The noise scheduling mechanism utilizes a DDIM scheduler configured with 100 training timesteps and \textit{squaredcos\_cap\_v2} beta schedule ranging from 0.0001 to 0.02. This configuration achieves stable training while enabling efficient inference with only 10 steps. This architecture design demonstrates particular strength in learning complex facial motion patterns and dynamic expressions. As validated by our experimental results, The progressive channel expansion provides sufficient capacity for hierarchical motion pattern extraction, while the FiLM conditioning mechanism enables effective integration of audio-visual context, resulting in more natural and expressive facial animations.

\begin{figure}[h]
\centering
\includegraphics[width=0.6\linewidth]{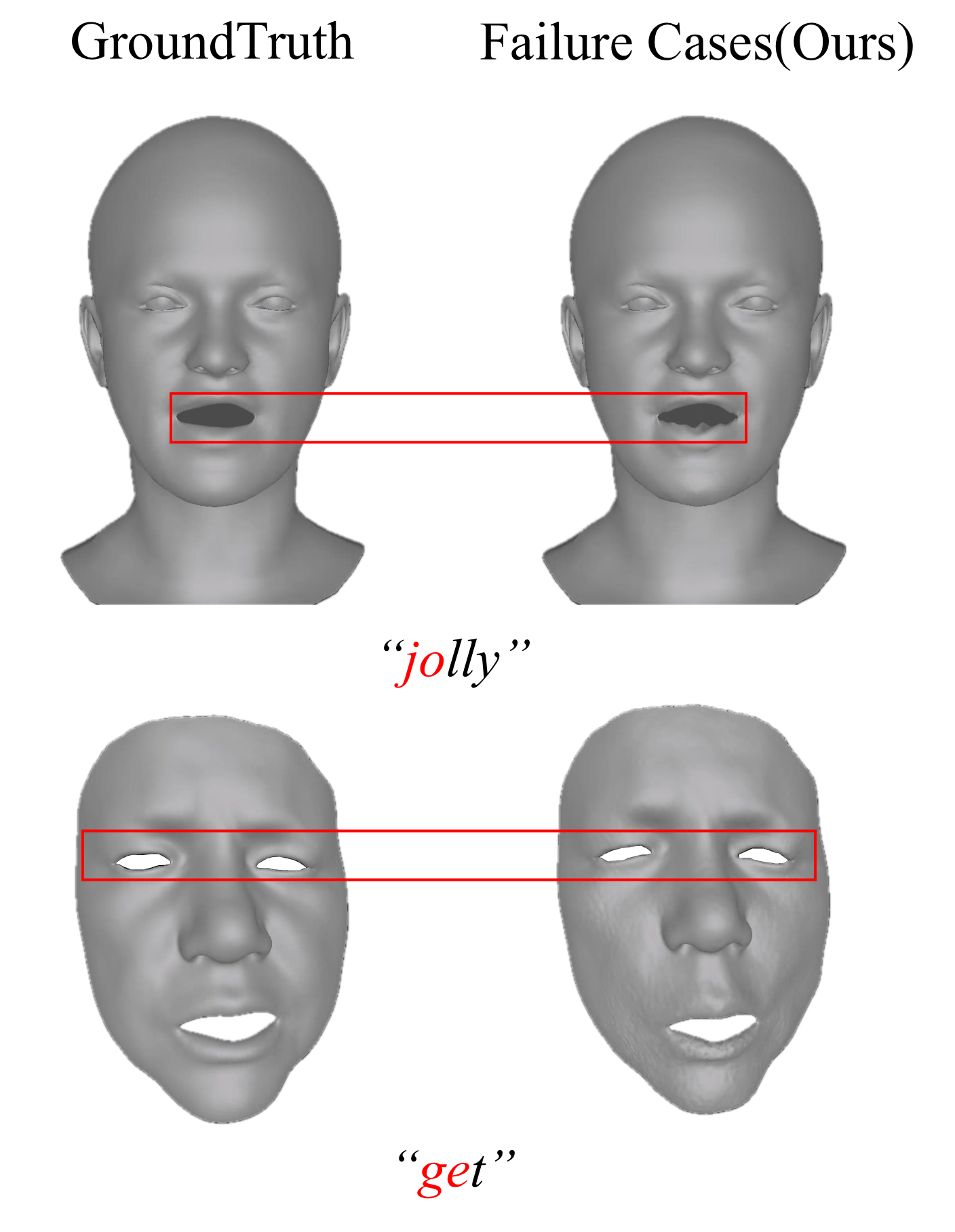}
\centering
\caption{Examples of the failure cases from the proposed approach. In extreme cases, the accumulated action predictions may lead to distortions in lip and eye shapes.}
\label{failure}
\end{figure}

\subsection{Failure Cases}
\label{sec:failure}
As shown in Fig. \ref{failure}, while our model demonstrates improved facial movements and expression generation overall, there are some specific scenarios that present opportunities for future improvement. In some extreme cases, our approach can show slight variations in lip shapes in the top row of Fig. \ref{failure} on the VOCASET dataset, and occasionally produces distorted eye regions in the bottom row of Fig. \ref{failure} on the BIWI dataset. These variations arise from our action prediction mechanism, where minor differences in action generation can accumulate during the predicting process. Specifically, when subtle variations occur during action prediction, the frame-by-frame vertices position updates may lead to slightly different facial features from the expected outcomes. This characteristic of policy-based action prediction mechanism, while enabling more dynamic and expressive animations in most cases, suggests potential areas for refinement in maintaining consistent facial feature integrity over a whole animation sequence. this challenge will be addressed in future work for balancing dynamic facial movements with anatomical consistency across animation sequences.

\subsection{User Study}
\label{sec:User_Study}
\begin{figure*}[h]
\centering
\includegraphics[width=\linewidth]{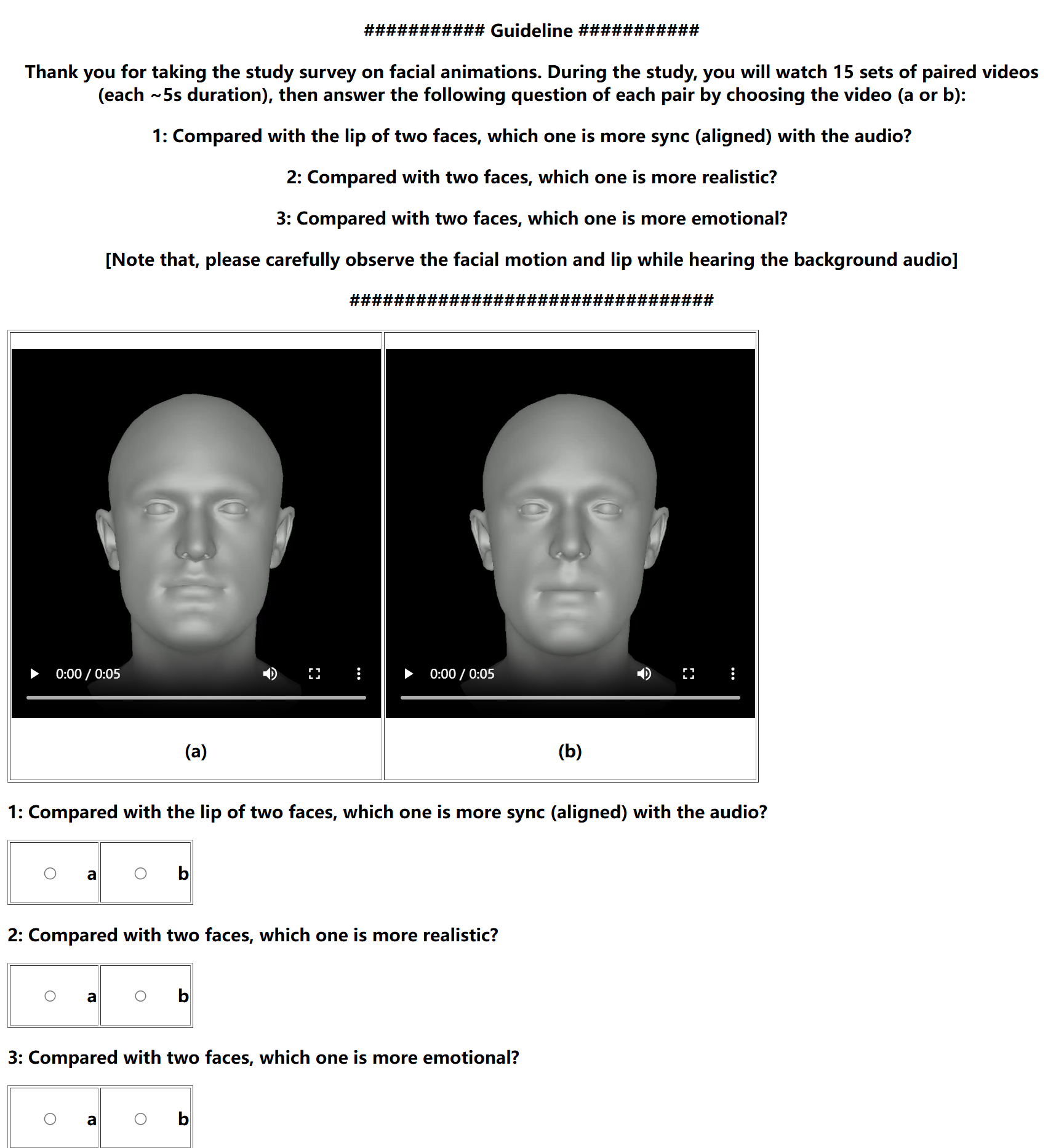}
\centering
\caption{Designed user study interface. Each participant need to answer 15 video pairs and here only one video pair is shown due to the page limit.}
\label{User_study_inter}
\end{figure*}

The designed user study interface is depicted in Fig. \ref{User_study_inter}. Participants are asked to make side-by-side comparisons and select the better animation based on their personal preferences. Each video pairs are selected from examples randomly and switch the side randomly between our model and competitors. Similar to multi-annotator learning frameworks that leverage annotator-specific reliability to improve subjective evaluation consistency \cite{SimLabel}, we adopt paired comparison and multi-participant voting to reduce individual bias in perceptual judgments. The completion time for each participant is about 15 minutes, with 15 video pairs and 3 questions for each pair.
The questions that participants are presented with are as follows: (1) Compared with the lip of two faces, which one is more sync (aligned) with the audio? (2) Compared with two faces, which one is more realistic? (3) Compared with two faces, which one is more emotional?

\clearpage

\end{document}